# STEREO ACOUSTIC PERCEPTION BASED ON REAL TIME VIDEO ACQUISITION FOR NAVIGATIONAL ASSISTANCE

Supreeth K Rao[1], Arpitha Prasad B[1], Anushree R Shetty[1], Chinmai[1],

R. Bhakthavathsalam[2], Rajeshwari Hegde[1]

[1]Dept. of Telecommunication Engineering, BMS College of Engineering, Bangalore.
supreethkrao@gmail.com, arpithaprasad@gmail.com,
anushree.shetty12@gmail.com, cpchinmai@gmail.com,
rajeshwari.hegde@gmail.com

[2] Supercomputer Education and Research Centre (SERC),
Indian Institute of Science, Bangalore, India.
bhaktha@serc.iisc.ernet.in

## ABSTRACT

*A smart navigation system (an Electronic Travel Aid) based on an object detection mechanism has been designed to detect the presence of obstacles that immediately impede the path, by means of real time video processing. This paper is discussed keeping in mind (however, not limited to) the navigation of the visually impaired. A video camera feeds images of the surroundings to a Da-Vinci Digital Media Processor, DM642, which works on the video, frame by frame. The processor carries out image processing techniques whose result contains information about the object in terms of image pixels. The algorithm aims to select the object which, among all others, poses maximum threat to the navigation. A database containing a total of three sounds is constructed. Hence, each image translates to a beep, where every beep informs the navigator of the obstacles directly in front of him. This paper implements an algorithm that is more efficient as compared to its predecessors.*

## 1. INTRODUCTION

Navigation of the blind involves many challenges such as walking on the road and avoiding obstacles. An Electronic Travel Aid (ETA) is a form of assistive technology having the purpose of enhancing mobility for the blind pedestrian [1]. Perhaps the most widely known device is the Laser-Cane or any other electronic cane [2], which might be a regular long cane with a built-in laser ranging system or a Geographic Information System (GIS). Nagarajan R, et al [3] have developed "NAVI: An Improved Object Identification for NAVI" which is a vision substitute system designed to assist blind people for autonomous navigation. Bin Ding, et al [4] proposed a blind navigation system based on Radio Frequency Identification (RFID), wireless and mobile communications technologies. The Mowat Sensor is an example of a pocket-sized device containing an ultrasonic air sonar system. When it detects an obstacle, the device vibrates, thereby signalling the user. Velazquez, et al [5] designed a prototype of the Intelligent Glasses, a novel non-invasive ETA to assist the visually impaired to navigate easily, safely and quickly among obstacles in indoor/outdoor 3D environments. Santosh S S, et al [6] proposed "BLI–NAV", blind navigation system exclusively designed for blind people which allows the blind people to travel through familiar and unfamiliar environment. Ran L, et al [7], proposed "Drishti: an integrated indoor/outdoor blind navigation system and service", which provides dynamic routing and rerouting ability through vocal prompts. Fernandes H, et al[8] proposed a paper that focused mainly in the development of a computer vision module for a Smart-Vision system. In research, the quest for designing a better ETA has always been a tough challenge [9].

Despite decades of effort, technology has not yet rewarded us with an electronic device that can completely complement or replace the long cane. There are many problems with currently available assistive devices. Firstly, the rangefinder technology is unreliable in its detection of step-downs or step-ups, such as curbs. Secondly, blind users find the tactile vibrations being used to encode the spatial information to be esoteric and difficult to interpret. In order to navigate, one must be able to detect "where" things are, in other words, the spatial positions of objects and other features in the physical world around us. Thus, we can reframe the problem of travelling as primarily a problem in spatial perception. Man is well endowed with a refined set of spatial sensing systems and they include binocular vision, binaural hearing and active touch [10], listed in the order of decreasing range of operation. To compensate for the loss of binocular vision, a blind traveller desires some sort of Display ETA that can spatially sense what is out there in his surroundings within a reasonable distance, and then convey this information in an easily interpretable format via the remaining senses of hearing and touch [11]. Many of those who are visually impaired can maintain their current employment or be trained for new work with the help of such aids.

This paper deals with a vision substitution system that is based on an image to sound conversion concept. It finds particular applications for the navigation of the visually impaired and even in the case of autonomous intelligent rovers. The output of this system can either be fed as an actuation to a smart control system, or converted to an audio signal which is fed to the blind person's earphones. This paper aims at creating a portable system that allows visually impaired individuals to travel through familiar and unfamiliar environments without the assistance of guides. This paper is organized as follows. Section II presents the proposed algorithm. Section III explains the methodology. Section IV deals with the hardware implementation. Section V deals with the results. The paper is concluded in section VI.

## 2. PROPOSED ALGORITHM

A vision acquisition device such as a video camera captures the information about the system surroundings. Image frames are procured from the video and subjected to a series of image processing techniques. Figure 1 shows the block diagram describing an overview of the algorithm used for the system implementation. The algorithm entrusts significant weightage to the location and size of the objects in the image under consideration. A flood function has been designed to calculate the same. These parameters are then used to assign priorities to the objects based on proximity and size. The object that has gained the highest priority is selected and depending upon its proximity and size, acoustic transformation is performed resulting in one out of two sounds. This sound is what gives the ultimate intended information about the surroundings that helps the user to have a collision free navigation.

Object recognition has been implemented to inform the user about what obstacles are present before him, rather than just their presence.

Figure1 shows the flowchart describing the series of image processing techniques applied on each image frame extracted from the incoming real time video.

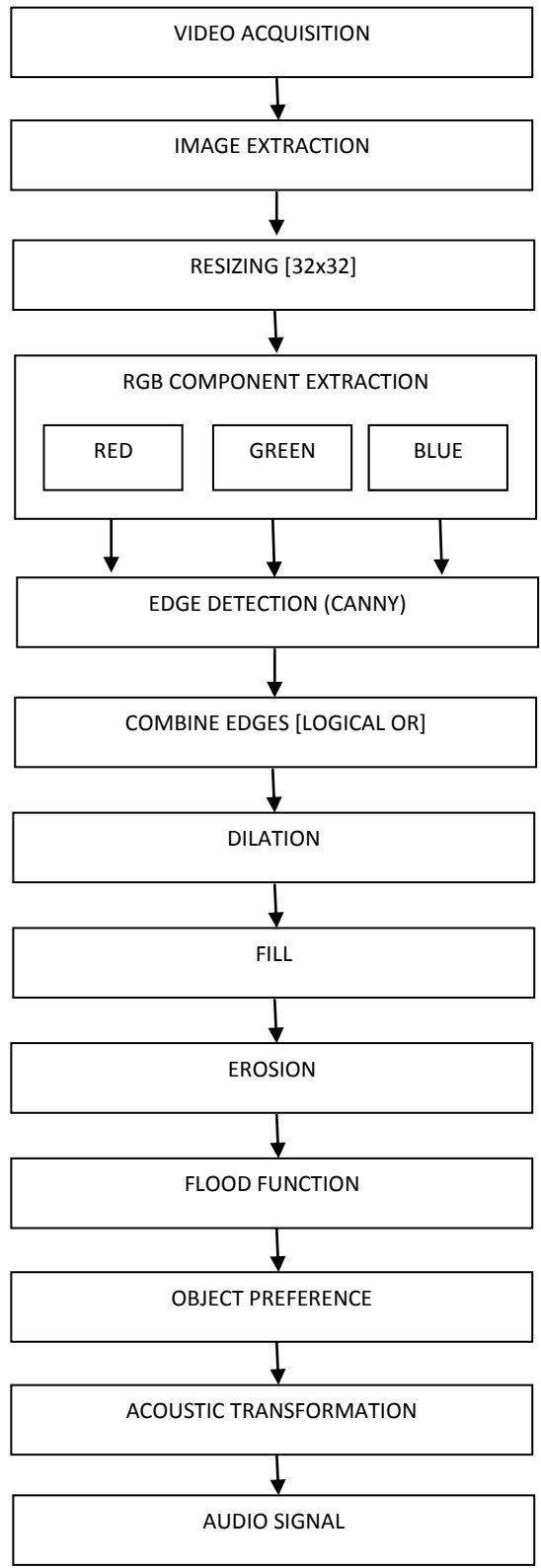

Figure1. Proposed Algorithm

## 3. METHODOLOGY

The proposed methodology makes use of the following standard pre-processing techniques:

### 3.1. Resizing

The image captured is resized to **32x32**. This is mainly to achieve the real time processing pre-requisite of small computational time, but it also provides flexibility of changing the camera (and thus the resolution). The above has been selected keeping in mind the data loss in resizing.

### 3.2. Edge detection

The objects present in an image are recognized by their boundaries. Edge detection is a technique that extracts the edges of the objects. There are various types of edge detection- sobel, canny, prewitt, laplacian to name a few, and the algorithm is friendly to both canny and sobel. The image that is acquired is a colour image consisting of red, green and blue (RGB) components [12]. Colour image segmentation offers greater accuracy when compared to edge detection in grayscale images [13]. Hence, we extract the RGB colour components of the resized image and perform edge detection on each. The results are combined by performing a logical OR operation on the three edge detected images to obtain a single binary image with clearly defined objects whose outline is in white.

### 3.3. Dilation

Morphological processing involves operations that process images based on shapes. They apply a structuring element to the input image that suitably altering it. Dilation and erosion are the two morphological operations that the algorithm uses.

Dilation has been used to connect broken edges in the edge detected image. A structural element of 2x3 'one's is appropriately chosen to dilate every white pixel in the binary image. This leads to thickening of the edges, hence connecting minor breaks.

### 3.4. Fill

As the size of the object has to be calculated, the area within each object must be obtained i.e., the number of pixels constituting the object. The image as of now consists of objects outlined in white. The fill function, when applied to the above image, fills in the black area within objects with white pixels. It can thus be seen that the object size is the number of white pixels it contains.

### 3.5. Erosion

Dilation adds pixels to the boundaries of objects in an image, while erosion removes pixels on object boundaries. While this sharpens the dilated objects, it also removes unwanted white specks, which would otherwise be viewed as small objects themselves. The structural element here is a disk with radius one.

### 3.6. Proximal Area

The location of the objects is an important aspect for collision free navigation. The presence of objects in certain parts of the image poses greater danger to the navigator than their presence in other parts. To understand this, let us divide an image into four parts, left, right, centre, back, as shown in figure 2.

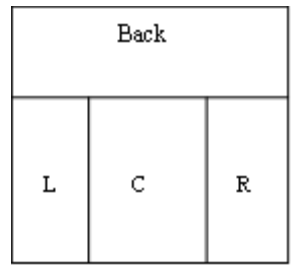

Figure2. Analysis of the image frame

Consider the following cases in Figure3 with the assumption that the user is walking straight ahead:

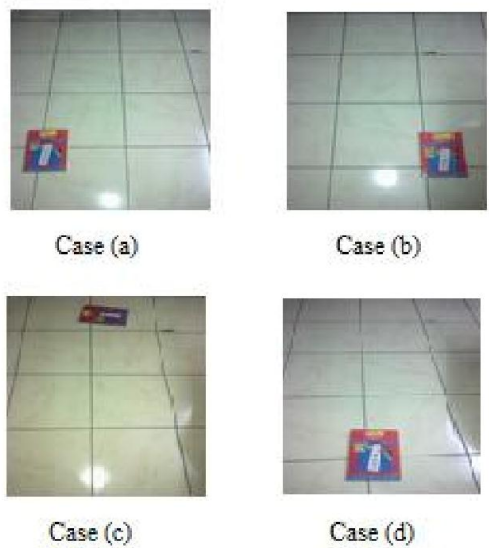

Figure3. Selection of Proximal Area

**Case (a):** An object is in the left (L) portion of the image and it is apparent that this object does not cause any obstruction to the navigator.

**Case (b):** The object in the right (R) portion of the image. This again will not obstruct the path of the person.

**Case (c):** Here, the object is at the back portion of the image. As we can see, the user can walk a short distance before he encounters this object.

**Case (d):** In this figure, however, where the object is present in the central portion of the image, we can see that the object causes an immediate obstruction to the navigator. Therefore, we can say that this region of the image should be given higher preference when compared to the other regions in the image. To classify the objects based on their location, the frame has been divided into three regions (Fig. 4), two of which constitute the high priority region – Proximal area consisting of A1 & A2 (Fig.2). Within the Proximal area, objects present in A1 are assigned the highest priority, while those present in A2 are assigned a lower priority.

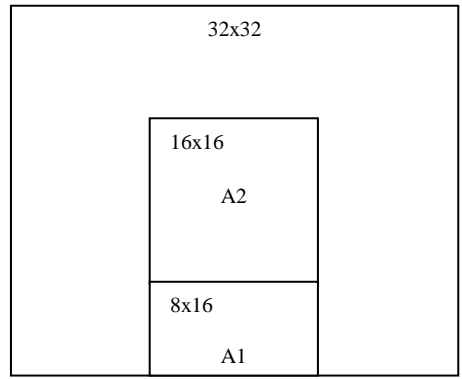

Figure4. Proximal Area

The preferences are assigned to A1 and A2 through the use of masks M1 and M2 consisting of ones. M1 is a mask of dimension 24x16 highlighting the objects in the areas A1 and A2. M2 is a mask of dimension 8x16 highlighting the objects in A1only. Using these masks, the following operations are performed:

M1 = Image & M1; (objects in A1 and A2 are highlighted)

M2 = Image & M2; (objects in A1 alone are highlighted)

Image = Image + M1 + M2,

This results in the image consisting of ones, twos and threes in the regions outside the proximal area, in A2 and in A1 respectively.

### 3.7. Flood Function

The objects in the image consist of ones, twos and threes; hence, calculating the object size would be to calculate the total number of ones, twos and threes. Also, the concentration of objects (which is the number of pixels) in A1 and A2 have to be calculated. This would mean counting the number of twos to calculate object concentration in A2, and threes to calculate the object concentration in A1. To calculate the size of the objects present in the frame, and its concentration in A1 and A2, a flood function is designed. The image is first searched for a one, two or three, and the flood function is called when these pixel intensities are encountered. The function called at a pixel spreads to its neighbours if its pixel intensities are not zeros. The function has been defined for a connectivity of eight, i.e, all eight neighbours of each pixel is checked for either a one, two or three. In effect, when a flood function is called at a pixel of an object, then it spreads to cover the entire object till it reaches the object's boundaries. During flooding, pixel count can be incremented (which would be the object size) and also, the number of twos and threes can be counted (which would mean the object concentration in A2 and A1 respectively). In order to prevent the function from flooding into already explored pixels, the intensities of those pixels where the function has been called are changed to zero. Therefore, not only does the function count the required values, it also shrinks the object into inexistence. This has no consequences as all necessary information has already been gathered. The scanning continues, and the procedure is repeated as and when other objects are encountered. A database containing object sizes and concentrations is thus created.

### 3.8. Priority Assignment

Objects falling in the Proximal Area should be given high priority. Not only should this been done, but importance should be given to the object size. Consider the following conflicting cases:
*Case 1*: If there are two objects lying in A2, the larger object should be given higher priority.
*Case 2*: If there is a large object in A2 and a smaller object in A1, as the user encounters the object in A1 first, objects in A1 should be given more priority than those in A2, regardless of the difference in size.
*Case 3*: If a small percentage of a huge object lies within A1and A2, and an object of smaller size lies only within A2, then that object whose concentration within A1 or A2 is more gains higher priority.

### 3.9. Acoustic Transformation

Each object by now would have gained a level of priority. Among all objects, that object which has gained the highest priority must have its presence conveyed to the blind user. Thus, the algorithm translates the object into an audio signal, resulting in a beep. All priority levels are categorized into three sounds: (i) a sound of high pitch indicating the presence of an object posing greatest threat and implying that the user must immediately change his direction (ii) a sound of medium pitch indicating that an object would soon pose the greatest threat if the user continues on his path and (iii) no sound, which informs that there is no object yet that can pose a threat.

## 4. HARDWARE IMPLEMENTATION

The system has used the Digital Media Processor TMS320DM642 (Version 3), which belongs to the Da Vinci family of Texas Instruments" C6000 series. The DM642 Evaluation Module (EVM) is a low-cost, high performance video imaging development platform designed to jump-start application development and evaluation of multi-channel, multi-format digital and other future proof applications. DM642 has been specially designed for real time video and audio processing, with dedicated video encoders and a decoder [14]. Leveraging the high performance of the TMS320C64x DSP core, this development platform supports TI's TMS320DM642, DM641 & DM640 digital media processors. The TMS320C64x™ DSPs (including the TMS320DM642 device) are the highest performance fixed-point DSP generation in the TMS320C6000™ DSP platform. The TMS320DM642 device is based on the second-generation high-performance, advanced VelociTI™ very-long-instruction-word (VLIW) architecture (VelociTI.2™) developed by Texas Instruments (TI), making these DSPs an excellent choice or digital media applications. DM642 offers a speed of 720MHz, 4 MBytes Flash, 32 MB of 133 MHz SDRAM and 256 kbit I2C EPROM [15]. The JTAG emulator used to communicate with the processor is XDS510USB Plus. A PAL (Phase Alternating Line) camera has been utilised to acquire the input (at 30 frames per second) to the system and a set of 3.5 mm jack ear-phones are used to provide the output of the system to the user [16]. Figure5 shows a schematic representation of the hardware implementation.

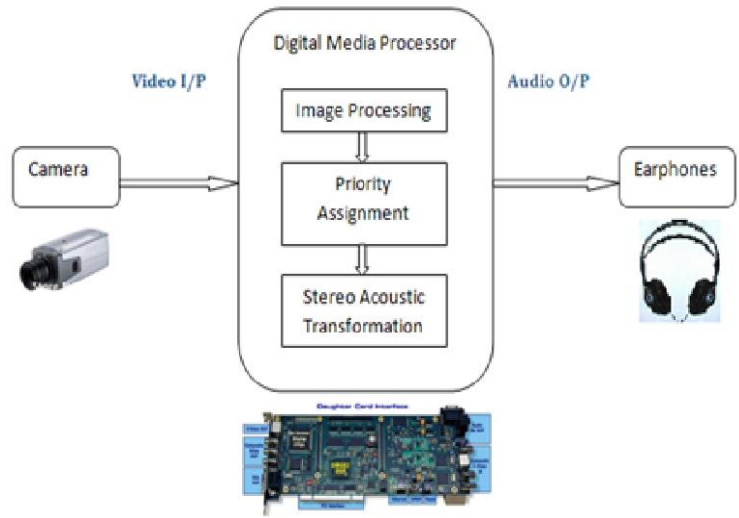

Figure5. Hardware Implementation using DM642

Figure 6 shows the experimental setup of the system.

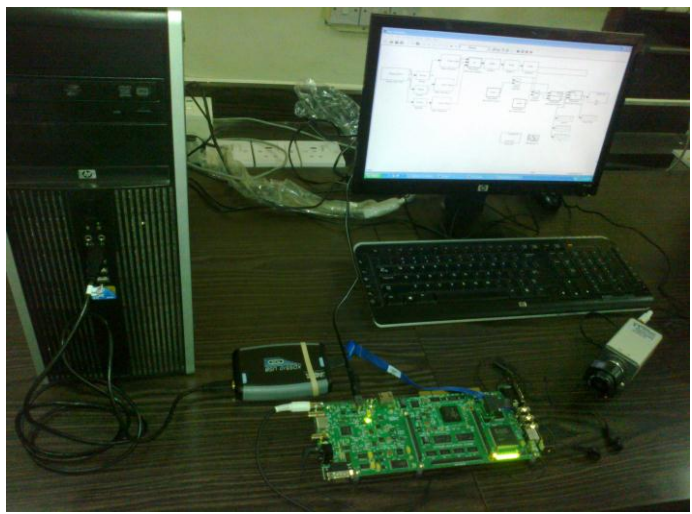

Figure 6. Experimental Setup

The video camera acquires video at real time and feeds the same to the processor (DM642). The Algorithm proposed detects the presence of obstacles, categorizes them based on their size and proximity. Stereo acoustic transformation is then performed to notify to the user about the presence of that object among others detected, which offers maximum threat to a collision free navigation.

## 5. RESULTS

This section shows the results for a sample image. The image has been resized as shown in Figure7 and the image processing techniques described above have been performed and the results of the same are shown in Figure8. The flood function then counts the object size and object concentration in A1 and A2. From Figure7, we infer that there are three objects that are important. It is evident from Table1 that the car to the bottom right gets the highest priority of 405 owing to its concentration in A1 (53) and size (113).

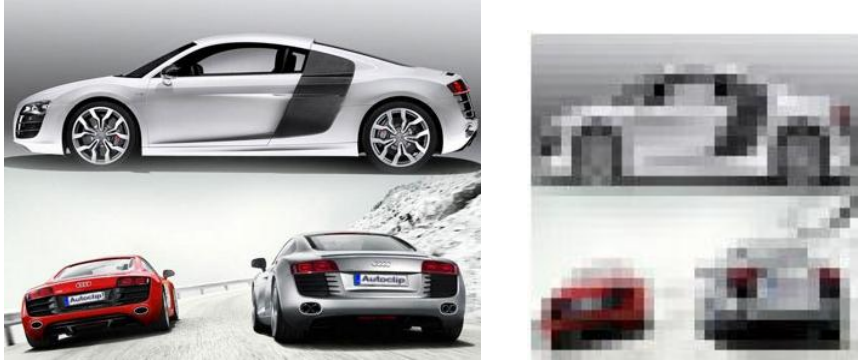

Figure 7. Image Extraction and Resizing

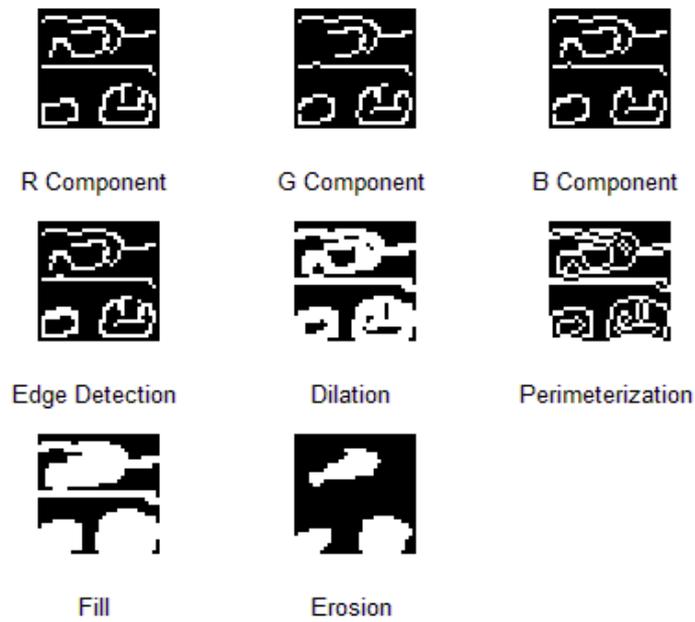

Figure 8. Image processing steps

Table 1. End results considering five objects

| A1 | 0 | 53 | 9 | 0 | 0 |
|---|---|---|---|---|---|
| A2 | 47 | 9 | 0 | 0 | 0 |
| Objsize | 100 | 113 | 59 | 0 | 0 |
| Priority | 45 | 405 | 55 | 0 | 0 |
| Highest Priority | 405 | | | | |

# 6. CONCLUSION

Acoustic vision is a sensory substitution system that acquires, processes, analyses, and interprets images from the real world and ultimately aims to provide a synthetic vision through sound, relevant to obstacle detection based navigation. Here, image extraction from a video input is carried out; boundaries of objects present in the image are identified and the objects' sizes are calculated. Importance is given to the size and proximity of objects through well-defined proximal areas. The flood function counts the size of objects and their concentration in these iris areas. Priority assignments then take place to categorize detected objects based on the amount of threat they pose and, finally, an acoustic transformation is performed to translate the visual information to a meaningful sound. This is executed by the Da Vinci Digital Media Processor, DM642 in half a second, and the real time system's execution returns again to the first step and the procedure is carried out repeatedly. Neuroscience and psychology research indicate recruitment of relevant brain areas in seeing with sound, as well as functional improvement through training. However, the extent to which cortical plasticity allows for functionally relevant rewiring or remapping of the human brain is still largely unknown and is being investigated in an open collaboration with research partners around the world. In effect, this system translates one kind of sensory perception into another, which the user eventually gets accustomed to owing to the neural plasticity of the human brain. The visually impaired people can now "see" by listening to the output of this algorithm. This proposed algorithm scores better as compared to its predecessor – NAVI: An Improved Object Identification for NAVI, due to the following facts:

Fuzzy Logic has been eliminated and replaced by a simple but yet an efficient thresholding mechanism. This makes the algorithm less computationally intensive.

A well-defined Proximal area divides the frame in such a way that the highest priority regions A1 and A2 are clearly identified, thus improving the accuracy of the obstacle detection.

An efficient Flood function has been developed to calculate the size of the detected objects in the least possible time.

The algorithm is implemented using a 720MHz processor thus greatly enhancing the speed at which the region before the user is interpreted by the system. This makes the system generic, as it can be deployed on a surveillance bot, wheel chair or a planetary exploration rover that is required to navigate in unfamiliar surroundings, with agility.

Implementation of object recognition and the inclusion of proximity sensors will augment the accuracy of the system to a considerable extent.

## ACKNOWLEDGEMENT

We are ever indebted to our college, BMS College of Engineering, for providing us with a healthy environment conducive to learning that enabled us to conceptualise and conduct a work such as this.

We would like to thank Dr. B Kanmani, Head of the department, Telecommunications Engineering, BMSCE, Bengaluru, under whose headship and encouragement, we were given access to a very well-equipped lab, without which this work would have just remained in a book of ideas.

We would like to express our sincere gratitude to Mr. Gowranga, SERC, Indian Institute of Science, Bengaluru, for his timely inputs that helped us progress towards our goal in the right direction.

Our deepest thanks to Mr. Jayaramudu, Cranes International Software Ltd., Bangalore, for his invaluable help in setting up the DSP DM642 Processor.

## Authors

**Supreeth K Rao** is currently pursuing his B.E at the Dept. of Telecommunications Engineering, BMS College of Engineering, Bangalore. He has successfully completed his final semester project at the Indian Institute of Science, Bangalore. He has a placement offer at IBM and a Research Assistantship at the Raman Research Institute, Bangalore. He aspires to pursue his M.S and PhD in Electrical Science. 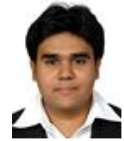

**Arpitha Prasad B** is currently pursuing her B.E at the Dept. of Telecommunications Engineering, BMS College of Engineering, Bangalore. She has successfully completed her final semester project at the Indian Institute of Science, Bangalore. She has a placement offer at TCS. She aspires to pursue her M.S and PhD in Mathematics. 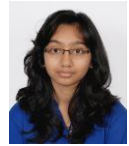

**Anushree R Shetty** is currently pursuing her B.E at the Dept. of Telecommunications Engineering, BMS College of Engineering, Bangalore She has successfully completed her final semester project at the Indian Institute of Science, Bangalore. She has a placement offer at Wipro Technologies. She aspires to pursue her MBA. 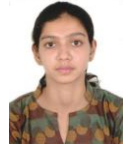

**Chinmai** is currently pursuing her B.E at the Dept.of Telecommunications Engineering, BMS College of Engineering, Bangalore. She has successfully completed her final semester project at the Indian Institute of Science, Bangalore. She has a placement offer at TCS. She aspires to pursue her M.Tech in Embedded Systems. 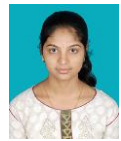

**Dr. R.Bhakthavathsalam** is presently working as a Senior Scientific Officer in SERC, IISc, Bangalore. His areas of interests are Pervasive Computing and Communication, Electromagnetics with a special reference to exterior differential forms. He was a fellow of Jawaharlal Nehru Centre for Advanced Scientific Research (1993-1995). He is a member of ACM & CSI. 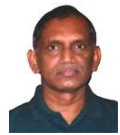

**Rajeshwari Hegde** is currently an Associate professor at the Dept. of Telecommunications Engineering, BMS College of Engineering, Bangalore. She received her M.E (Electronics) from BMS College of Engineering, Bangalore and B.E from National Institute of Engineering, Mysore. She has done her PhD under the guidance of Dr K S Gurumurthy at UVCE, Bangalore University. She has published 43 research papers in international conferences, national conferences and reputed journals. 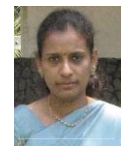